\documentclass[letterpaper]{article} 
\usepackage[submission]{aaai24}  
\usepackage{times}  
\usepackage{helvet}  
\usepackage{courier}  
\usepackage[hyphens]{url}  
\usepackage{graphicx} 
\usepackage{natbib}  
\usepackage{caption} 
\usepackage{algorithm}
\usepackage{algorithmic}

\usepackage{xcolor}         
\usepackage{color}
\usepackage{amsmath}

\usepackage{multirow}

\usepackage{graphicx}
\usepackage{subfigure}
\usepackage[capitalize]{cleveref}

\usepackage{amssymb}

\usepackage{scalerel}

\newcommand{\tabincell}[2]{\begin{tabular}{@{}#1@{}}#2\end{tabular}}

\usepackage{newfloat}
\usepackage{listings}
\DeclareCaptionStyle{ruled}{labelfont=normalfont,labelsep=colon,strut=off} 
\floatstyle{ruled}
\newfloat{listing}{tb}{lst}{}
\floatname{listing}{Listing}
\title{WaveAttack: Asymmetric Frequency Obfuscation-based Backdoor Attacks Against Deep Neural Networks}
\author{
    Jun Xia\textsuperscript{\rm 1, 2*},
    Zhihao Yue\textsuperscript{\rm 1*},
    Yingbo Zhou\textsuperscript{\rm 1},
    Zhiwei Ling\textsuperscript{\rm 1},
    Xian Wei\textsuperscript{\rm 1}
    Mingsong Chen\textsuperscript{\rm 1}
}
\affiliations{
    \textsuperscript{\rm 1}Software Engineering Institute, East China Normal University\\
     \textsuperscript{\rm 2}Computer Science and Engineering, University of Notre Dame \\
        jxia4@nd.edu,  mschen@sei.ecnu.edu.cn

}

\usepackage{bibentry}

\begin{document}

\maketitle

\begin{abstract}
Due to the increasing popularity of Artificial Intelligence (AI), more and more backdoor attacks are designed to mislead Deep Neural Network (DNN) predictions by manipulating training samples and training processes. 
Although backdoor attacks have been investigated in various real scenarios, they still suffer from the problems of both low fidelity of poisoned samples and non-negligible transfer in latent space, which make them easily detectable by existing backdoor detection algorithms.
To overcome this weakness, we propose a novel frequency-based backdoor attack method named WaveAttack, which obtains image high-frequency features through Discrete Wavelet Transform (DWT) to generate backdoor triggers.
Furthermore, we introduce an asymmetric frequency obfuscation method, which can add an adaptive residual in the training and inference stage to improve the impact of triggers and further enhance the effectiveness of WaveAttack.
Comprehensive experimental results show that WaveAttack not only achieves higher stealthiness and effectiveness, but also outperforms state-of-the-art (SOTA) backdoor attack methods in the fidelity of images by up to 28.27\% improvement in PSNR, 1.61\% improvement in SSIM, and 70.59\% reduction in IS. 
\end{abstract}


\section{Introduction}
Along with the prosperity of Artificial Intelligence (AI), Deep Neural Networks (DNNs) have become increasingly prevalent in numerous safety-critical domains for precise perception and real-time control, such as autonomous vehicles \cite{autonomous_driving}, medical diagnosis \cite{medical}, and industrial automation \cite{industrial_automation}. 
However, the trustworthiness of DNNs faces significant threats due to various 
notorious adversarial and backdoor attacks.
Typically, adversarial attacks \cite{Carlini017} manipulate input data during the inference stage to induce incorrect predictions by a trained DNN, whereas backdoor attacks \cite{badnets} tamper with training samples or training processes to embed concealed triggers during the training stage, which can be exploited to generate malicious outputs. 
Although adversarial attacks on neural networks frequently appear in various scenarios,  backdoor attacks have attracted more attention due to their stealthiness and effectiveness. Generally, 
the performance of   backdoor attacks can be evaluated by the following three objectives of an adversary: 
i) {\it efficacy} that refers to the effectiveness of an attack in causing the target model to produce incorrect outputs or exhibit unintended behavior; 
ii) {\it specificity} that denotes the precision of the attack in targeting a specific class; and iii) {\it fidelity} that represents the degree to which adversarial examples or poisoned training samples are indistinguishable from their benign counterparts \cite{attack-objective}. 
Note that efficacy and specificity represent the effectiveness of backdoor attacks, while fidelity denotes the stealthiness of backdoor attacks.

Aiming at higher stealthiness and effectiveness, existing 
backdoor attack methods (e.g., IAD \cite{backdoorat-IAD}, WaNet \cite{backdoorat-WaNet}, BppAttack \cite{backdoorat-Bpp}, and FTrojan \cite{backdoorat-FTrojan}) are built  based on
various optimizations, which can be mainly classified
into two categories. 
 The first one is the \textit{sample minimal impact} methods that can optimize the size of the trigger and minimize its pixel value, making the backdoor trigger hard to detect in training samples for the purpose of achieving a high stealthiness of a backdoor attacker.
Although these methods are promising in backdoor attacks, due to the explicit trigger influence on training samples, they cannot fully evade the existing backdoor detection methods based on training samples. 
The second one is the  \textit{latent space obfuscation-based} methods, which can be integrated into any existing backdoor attack methods. By employing asymmetric samples, these methods can obfuscate the latent space between benign samples and poisoned samples \cite{backdoorat-adapt, xia2022enhancing}.
Although these methods can bypass latent space detection techniques, they greatly suffer from low
image quality, making them extremely difficult to apply in practice. 
%
Therefore, \textit{how to improve both the effectiveness and stealthiness of backdoor attacks while minimally impacting the  quality of training samples is becoming a significant challenge in the development of backdoor attacks, especially when facing various state-of-the-art backdoor detection methods}.



This paper draws inspiration from the work \cite{DBLP:conf/cvpr/WangWHX20} where Wang et al. find that high-frequency features can enhance the generalization ability of DNNs and remain imperceptible to humans. To acquire high-frequency components (i.e., high-frequency features), wavelet transform has been widely investigated in various image-processing tasks \cite{DBLP:conf/cvpr/LiSGL20, DBLP:conf/iccv/YuZLPMXM21, DBLP:conf/nips/ZhongSYLZ18}. This paper introduces a novel frequency-based backdoor attack method named WaveAttack, which utilizes Discrete Wavelet Transform (DWT)  to extract the high-frequency component for backdoor trigger generation.  
Furthermore, to improve the impact of triggers and further enhance the effectiveness of our approach, we employ \textit{asymmetric frequency obfuscation} that utilizes an asymmetric coefficient of the trigger in the high-frequency domain during the training and inference stages. 
This paper makes the following three contributions:
\begin{itemize}

\item  
We introduce a frequency-based backdoor trigger generation method named WaveAttack, which can effectively generate the backdoor residuals for the high-frequency component based on DWT, thus ensuring the high fidelity of poisoned samples.

\item  We propose a novel asymmetric frequency-based obfuscation backdoor attack method to enhance its stealthiness and effectiveness, which can not only increase stealthiness in the latent space but also improve the Attack Success Rate (ASR) in training samples.

\item We conduct comprehensive experiments to demonstrate that  WaveAttack outperforms  SOTA backdoor attack methods regarding both stealthiness and effectiveness.  

\end{itemize}




\section{Related Work}\label{rel_work}
\textbf{Backdoor Attack.}
Typically, backdoor attacks try to embed backdoors into  DNNs by manipulating their input
samples and training processes. In this way, adversaries can control DNN outputs through concealed triggers, thus resulting in manipulated predictions \cite{li2020backdoor}.
Based on whether the training process is manipulated, existing backdoor attacks can be categorized into two types, i.e.,  \textit{training-unmanipulated}  and \textit{training-manipulated} attacks.
Specifically, the training-unmanipulated attacks only inject
a visible or invisible trigger into the training samples of some DNN, leading to its recognition errors
\cite{badnets}. 
For instance, Chen et al. \cite{physics_backdoor} introduced a Blend attack that generates poisoned data by merging benign training samples with specific key visible triggers.
Moreover, there exist a large number of invisible trigger-based backdoor attack methods, such as
natural reflection \cite{backdoorat-reflect}, human imperceptible noise \cite{zhong2020backdoor}, and image perturbation \cite{backdoorat-FTrojan}, which exploit the changes induced by real-world physical environments.
 %
Although these training-unmanipulated
attacks are promising, due to their substantial impacts on training sample quality,  most of them still can be  easily
identified somehow.
%
%
As an alternative, the training-manipulated  attacks \cite{backdoorat-WaNet, backdoorat-Bpp}
assume that adversaries
from some malicious third party can control the key steps of the training process, thus achieving
a  stealthier attack.
Although the above two categories of backdoor attacks
are promising, most of them struggle with the coarse-grained optimization of effectiveness and stealthiness,  complicating the acquisition of superior backdoor triggers. Due to the significant difference in latent space and low poisoned sample fidelity, they  cannot  evade the latest backdoor detection methods.

\textbf{Backdoor Defense.}
There are two major types of backdoor defense methods, i.e., 
the {\it detection-based defense} and {\it erasure-based defense}. The detection-based defenses can be further classified into two categories, i.e.,  sample-based  and latent space-based detection methods. Specifically, sample-based detection methods can identify the distribution differences between poisoned samples and benign samples \cite{DBLP:conf/acsac/GaoXW0RN19, backdoordf-imgdet-1, DBLP:conf/eccv/DoHLNTRNSV22}, while
latent space-based detection methods aim to find the disparity between the latent spaces of poisoned samples and benign samples \cite{tran2018spectral, hayase2021spectre}. 
%
Unlike the above detection strategies that aim to   prevent the injection of backdoors into DNNs by identifying poisoned samples during the training stages, the erasure-based defenses
can  eradicate the backdoors from DNNs. So far, the 
 erasure-based defenses
can  be classified  into three categories, i.e.,  poison suppression-based, model reconstruction-based, and trigger generation-based defenses. 
The poison suppression-based methods \cite{defense-abl} utilize the differential learning speed between poisoned and benign samples during training to mitigate the influence of backdoor triggers on DNNs. The model reconstruction-based methods \cite{liu2018fine, xia2022ARGD} leverage a selected set of benign data to rebuild  DNN models, aiming to mitigate
the impact of backdoor triggers. 
The trigger generation-based methods \cite{NC} reverse-engineer  backdoor triggers by capitalizing on the effects of backdoor attacks on training samples.

%
%
To the best of our knowledge, WaveAttack is the first attempt to generate triggers for the high-frequency component obtained through DWT.
Unlike existing backdoor attack methods, WaveAttack considers both the fidelity of poisoned samples and latent space obfuscation.
By using asymmetric frequency obfuscation, WaveAttack can  not only acquire backdoor attack effectiveness but also achieve high stealthiness regarding both image quality and latent space.


\section{Our Method}\label{approach}

In this section, we first present the preliminaries and the threat model. Then, we show our motivations for adding triggers to the high-frequency components. Finally, we detail the attack process of our method WaveAttack. 

\begin{figure*}[ht]
\vspace{-0.15in}
  \centering
  \subfigure[Original\label{motivation:original}]{\includegraphics[width=1.05in]{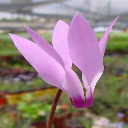}}
  \subfigure[LL with noises\label{motivation:LL}]{\includegraphics[width=1.05in]{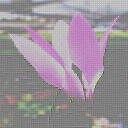}}
   \subfigure[LH with noises\label{motivation:LH}]{\includegraphics[width=1.05in]{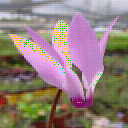}} 
   \subfigure[HL with noises\label{motivation:HL}]{\includegraphics[width=1.05in]{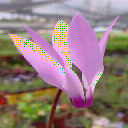}} 
   \subfigure[HH with noises\label{motivation:HH}]{\includegraphics[width=1.05in]{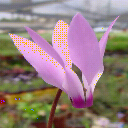}} \\
   \vspace{-0.10in}
  \caption{A motivating example for the  
  backdoor trigger design on high-frequency components.
  } \label{motivation}
\vspace{-0.10in}
\end{figure*}

\subsection{Preliminaries}

\textbf{Notations.}
We follow the training scheme of Adapt-Blend \cite{backdoorat-adapt}. Let $\mathcal{D}=\left\{(\boldsymbol{x}_i,y_i)\right\}_{i=1}^N$ be a clean training dataset, where $\boldsymbol{x}_i\in \mathbb{X}=\{0,1,...,255\}^{C \times W \times H}$ is an image, and $y_i\in \mathbb{Y}=\{1,2,..., K\}$ is its corresponding 
 label. Note that $K$ represents the number of labels. For a given training dataset, we select a subset of $\mathcal{D}$ with a poisoning rate $p_{a}$ as the \textit{payload samples} $\mathcal{D}_a = \{(\boldsymbol{x'}_i,y_t) | \boldsymbol{x'}_i=T(\boldsymbol{x}_i), \boldsymbol{x}_i\in \mathbb{X}\}$, where $T(\cdot)$ is a backdoor transformation function, and  $y_t$ is an adversary-specified target label. We use a subset of $\mathcal{D}$ with poisoning rate $p_{r}$ as the \textit{regularization samples} $\mathcal{D}_r = \{(\boldsymbol{x'}_i,y_i) | \boldsymbol{x'}_i=T(\boldsymbol{x}_i), \boldsymbol{x}_i\in \mathbb{X}\}$. For a given dataset, a backdoor attack adversary tries to train a backdoored model $f$ that predicts $\boldsymbol{x}$ as its corresponding label, where $\boldsymbol{x} \in \mathcal{D} \cup \mathcal{D}_a \cup \mathcal{D}_r$.

\textbf{Threat Model.}
Similar to existing backdoor attack methods \cite{backdoorat-IAD, backdoorat-WaNet, backdoorat-Bpp}, we assume that adversaries have complete control over the training datasets, the training process, and model implementation. 
They can embed backdoors into the DNNs by poisoning the given training dataset. Moreover, in the inference stage, we assume that adversaries can only query backdoored models using any samples.

\textbf{Adversarial Goal.}
Throughout the attack process, adversaries strive to meet two core goals, i.e., effectiveness and stealthiness. Effectiveness indicates that adversaries try to train backdoored models with a high ASR  while ensuring that the decrease in Benign Accuracy (BA) remains imperceptible.
Stealthiness denotes that samples with triggers have high fidelity, and there is no latent separation between poisoned and clean samples in the latent space.

\subsection{Motivation}
Unlike humans that are not sensitive to high-frequency features, DNNs can effectively
 learn high-frequency features of images \cite{DBLP:conf/cvpr/WangWHX20}, which can be used 
 for the purpose of backdoor trigger generation. In other words, the poisoned samples generated by high-frequency features can easily 
 escape from various examination methods by humans.
Based on this observation, if we can design 
backdoor triggers on top of high-frequency features, the stealthiness of corresponding backdoored attacks can be ensured.
%
To obtain high-frequency components from training samples, we resort to Discrete Wavelet Transform (DWT) to capture characteristics from both time and frequency domains \cite{shensa1992discrete}, which enables the extraction of multiple frequency components from training samples. 
The reason why we adopt DWT rather than Discrete Cosine Transform (DCT) is  that DWT can better capture high-frequency features from training samples (i.e., edges and textures) and allow superior reverse operations during both encoding and decoding phases, thus minimizing the impact on the fidelity of poisoned samples.
 %
In our approach, we adopt a classic and effective biorthogonal wavelet transform method (i.e., Haar wavelet \cite{daubechies1990wavelet}), which mainly contains four kernels operations, i.e., $LL^T$, $LH^T$, $HL^T$, and $HH^T$. 
Here $L$ and  $H$ denote the low and high pass filters, respectively, where 
    $L^T=\frac{1}{\sqrt{2}}\begin{bmatrix}1 \ \ 1 \end{bmatrix}, 
    H^T=\frac{1}{\sqrt{2}}\begin{bmatrix}-1 \ \ 1 \end{bmatrix}$.
 Note that, based on the four operations, the Haar wavelet can decompose an image into four frequency components (i.e., $LL$, $LH$, $HL$, $HH$) using DWT, where $HH$ only contains the high-frequency information of a  sample.
 Meanwhile, the Haar wavelet can reconstruct the image from the four frequency components via the Inverse Discrete Wavelet Transform (IDWT). 
 To verify the motivation of our approach, 
 Figure \ref{motivation} 
 illustrates the impact of adding the same  noises to different frequency components on an image, i.e., Figure \labelcref{motivation:original}. 
We can find that, compared with the other three poisoned images, i.e., Figure \labelcref{motivation:LL,motivation:LH,motivation:HL}, it is much more difficult to figure out 
the difference between the original image and the 
poisoned counterpart on HH, i.e., Figure \labelcref{motivation:HH}. 
Therefore, it is more suitable to inject triggers into the high-frequency component (i.e., HH) for the backdoor attack purpose.


\subsection{Implementation of WaveAttack}
\begin{figure}[ht]
\vspace{-0.1in}
\centering
\includegraphics[width=3.3in]{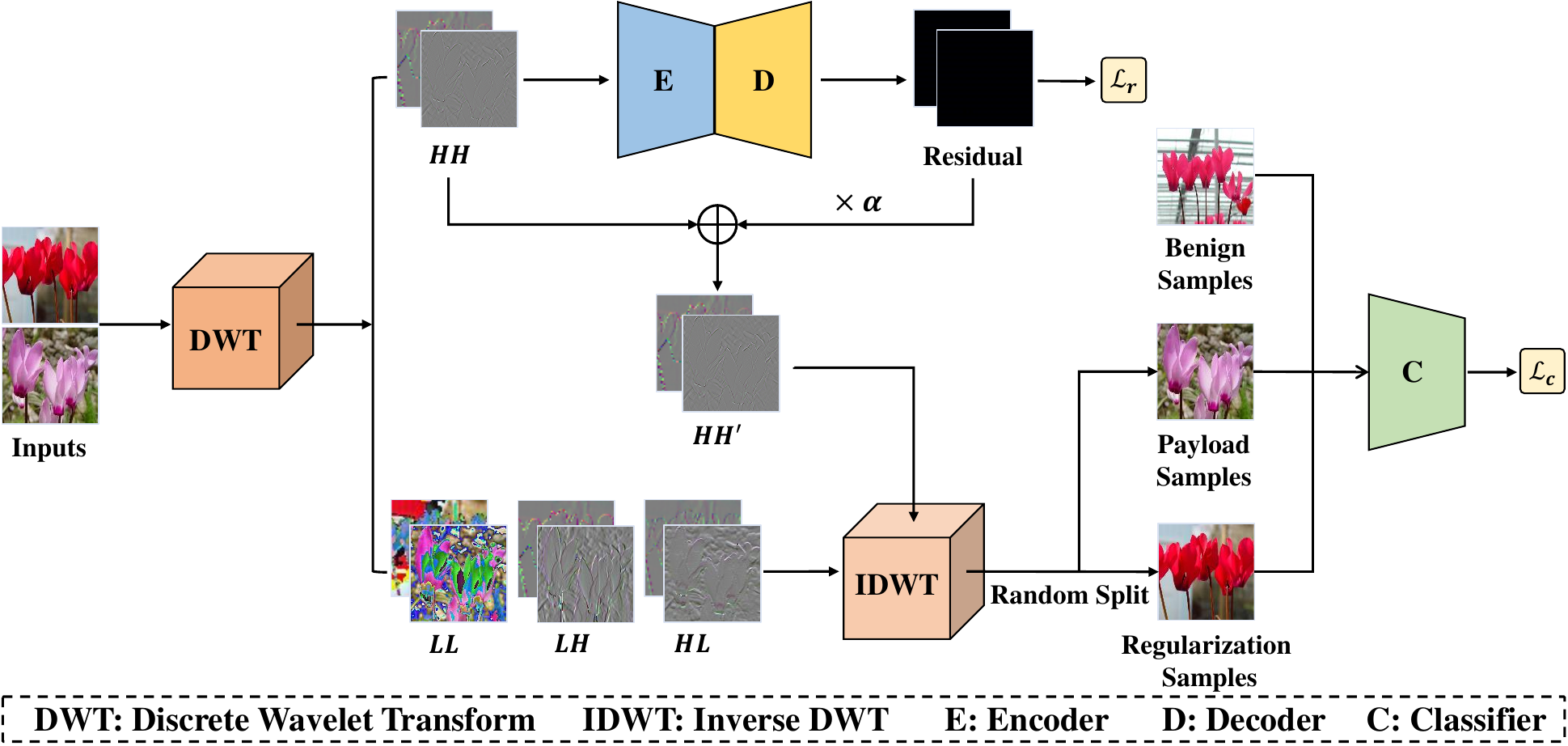}
\caption{Overview of our attack method WaveAttack.} \label{figure_overview}
\vspace{-0.1in}
\end{figure}

In this subsection, we introduce the design of our WaveAttack approach. 
Figure \ref{figure_overview} shows the overview of WaveAttack. 
We first poisoned samples through our trigger design to construct the poisoned samples, which contain payload samples and regularization samples. 
Then, we use benign samples, payload samples, and regularization samples to train a classifier to achieve the core goals of WaveAttack. 

\textbf{Trigger Design.}
As aforementioned, our WaveAttack approach aims to achieve a stealthier backdoor attack, introducing triggers into the $HH$ frequency component. Figure \ref{figure_overview} contains the process of generating triggers by WaveAttack. First, we obtain the $HH$ component of samples through DWT. Then, to generate imperceptible and sample-specific triggers, we employ an encoder-decoder network as a generator $g$. These generated triggers are imperceptible additive residuals. Moreover, to achieve our asymmetric frequency obfuscation, we multiply the residuals by a coefficient $\alpha$. 
We can generate the poisoned $HH'$ component with the triggers as follows:
 \begin{equation}    \label{eq_trigger}
    \boldsymbol{HH'}= \boldsymbol{HH} + \alpha \cdot g(\boldsymbol{HH}; \boldsymbol{\omega}_g),
\end{equation}
where $\boldsymbol{\omega}_g$ is the generator parameters. Finally, we can utilize IDWT to reconstruct four frequency components of poisoned samples. Specifically, we use a U-Net-like \cite{UNet} generator to obtain residuals, though other methods, such as VAE \cite{VAE}, can also be used by the adversary. This is because the skip connections of U-Net can effectively preserve the features of inputs with minimal impacts \cite{UNet}.

\textbf{Optimization Objective.} Our attack method WaveAttack has two networks to optimize. We aim to optimize a generator $g$ to generate small residuals with a minimal impact on samples. 
Furthermore, we aim to optimize a backdoored classifier $f$, which can enable the effectiveness and stealthiness of WaveAttack. 
For the first optimization objective, we use the $L_\infty$ norm to optimize small residuals. 
The optimization objective is defined as follows:
 \begin{equation}    \label{eq_l_infty}
     \mathcal{L}_{r}= ||g(HH;\boldsymbol{\omega}_g)||_\infty.
\end{equation}
As for the second optimization objective, we train the classifier by the cross-entropy loss function in $\mathcal{D}$, $\mathcal{D}_a$, and $\mathcal{D}_r$ dataset. 
The optimization objective is defined as follows:
 \begin{equation}    \label{eq_l_classifier}
     \mathcal{L}_{c}= 
     \mathcal{L}(\boldsymbol{x}_{p},y_t;\boldsymbol{\omega}_c)
     + \mathcal{L}(\boldsymbol{x}_{r},\boldsymbol{y};\boldsymbol{\omega}_c)
    + \mathcal{L}(\boldsymbol{x}_{b},\boldsymbol{y};\boldsymbol{\omega}_c),
\end{equation}
where $\mathcal{L}(\cdot)$ is the cross-entropy loss function, $\boldsymbol{\omega}_c$ is the classifier parameters, $\boldsymbol{x}_{b}\in\mathcal{D}, \boldsymbol{x}_{p}\in\mathcal{D}_a$, and $\boldsymbol{x}_{r}\in\mathcal{D}_r$. The total loss function is as follows:
 \begin{equation}    \label{eq_l_total}
     \mathcal{L}_{total}= \mathcal{L}_{c} + \mathcal{L}_{r}.
\end{equation}

\begin{algorithm}[H]
    \caption{Training of WaveAttack}
    \label{alg_waveattack}
\textbf{Input}:
            i) $\mathcal{D}$, benign training dataset; 
            ii) $\boldsymbol{\omega_{g}}$, randomly initialized generator parameters; 
            iii) $\boldsymbol{\omega_{f}}$, randomly initialized classifier parameters; 
            iv) $p_a$, rate of payload samples. 
            v) $p_r$, rate of regularization samples.
            vii) $y_t$, target label.
            vi) $E$, \# of epochs in training process. \\
\textbf{Output}: 
 i) $\boldsymbol{\omega}_g$,  well-trained generator model, 
 ii) $\boldsymbol{\omega}_c$,  well-trained classifier model.\\
\textbf{WaveAttack Training:}

\begin{algorithmic}[1] 
\FOR{$e=1,\dots,E$}\label{line:trainStart}
    \FOR{$(\boldsymbol{x},\boldsymbol{y})$ in $\mathcal{D}$}\label{line:loadData}
        \STATE $b$ $\leftarrow$ $\boldsymbol{x}$.shape[0] \\
        \STATE $n_{m}$ $\leftarrow$ $(p_a+p_r)\times b$ \\
         \STATE $n_{a}$ $\leftarrow$ $p_a\times b$ \\
         \STATE $n_{r}$ $\leftarrow$ $p_r\times b$ \\
        \STATE $\boldsymbol{x}_{m}$ $\leftarrow$ $\boldsymbol{x}$[:$n_{m}$] \\
         \STATE $\boldsymbol{LL},\boldsymbol{LH},\boldsymbol{HL},\boldsymbol{HH}$ $\leftarrow$ $DWT(\boldsymbol{x}_{m})$ \\
         \STATE $\boldsymbol{resdiual}$ $\leftarrow$ $\alpha \cdot g(\boldsymbol{HH};\boldsymbol{\omega}_g)$ \\
         \STATE $\boldsymbol{HH}'$ $\leftarrow$ $\boldsymbol{HH}+\boldsymbol{resdiual}$ \\
         \STATE $\boldsymbol{x}_{m}$ $\leftarrow$ $IDWT(\boldsymbol{LL},\boldsymbol{LH},\boldsymbol{HL},\boldsymbol{HH}')$ \\
         \STATE $\mathcal{L}_{1}$ $\leftarrow$ $\mathcal{L}(\boldsymbol{x}_m$[$n_a$:],$y_t$;$\boldsymbol{\omega}_c)$ \\
         \STATE $\mathcal{L}_{2}$ $\leftarrow$ $\mathcal{L}(\boldsymbol{x}_m$[:$n_r$],$\boldsymbol{y}$[$n_{a}$:$n_{r}$];$\boldsymbol{\omega}_c)$ \\
         \STATE $\mathcal{L}_{3}$ $\leftarrow$ $\mathcal{L}(\boldsymbol{x}$[$n_{m}$:],$\boldsymbol{y}$[$n_{m}$:];$\boldsymbol{\omega}_c)$ \\
         \STATE $\mathcal{L}$ $\leftarrow$ $\mathcal{L}_1+\mathcal{L}_2+\mathcal{L}_3+||\boldsymbol{resdiual}||_\infty$ \\
        \STATE $\mathcal{L}$.backward() \\
        \STATE update($\boldsymbol{\omega}_g,\boldsymbol{\omega}_c$) \\
    \ENDFOR
\ENDFOR
\STATE \textbf{Return} $\boldsymbol{\omega}_g,\boldsymbol{\omega}_{c}$

\end{algorithmic}
\vspace{-0.01in}
\end{algorithm}

\textbf{Algorithm Description.} Algorithm \ref{alg_waveattack} details the training process of our WaveAttack approach. 
At the beginning of WaveAttack training (Line $2$), the adversary randomly selects a minibatch data $(\boldsymbol{x},\boldsymbol{y})$ from $\mathcal{D}$, which has $b$ training samples. 
Lines $4$-$6$ calculate the number of poisoned samples, payload samples, and regulation samples, respectively. Lines $7$-$11$ denote the process of modifying samples by injecting triggers into the high-frequency component. After acquiring the modified samples in Line $7$, Line $8$ decomposes the samples into four frequency components by DWT. Then, in Lines $9$-$10$, we add the residual to the HH frequency component by Equation \ref{eq_trigger}. Line $11$ reconstructs the samples from the four frequency components via IDWT. Lines $12$-$15$ compute the optimization object by Equations~\labelcref{eq_l_infty,eq_l_classifier,eq_l_total}. In Lines $16$-$17$, we can use an optimizer (e.g.,  SGD optimizer) to update the parameters of the generator model and classifier model. Line $20$ returns the well-trained generator model parameters $\boldsymbol{\omega}_g$ and the classifier model parameters $\boldsymbol{\omega}_c$.

 \textbf{Asymmetric Frequency Obfuscation.} According to the work in  \cite{backdoorat-adapt}, regularization samples $\mathcal{D}_r$ can make DNNs learn the semantic feature of each class and the trigger feature, which can make the backdoor attack stealthy in the latent space. However, using the same trigger during the inference process may diminish the effectiveness of the attack method. 
 Hence, it is crucial to devise an asymmetric frequency obfuscation method to enhance the effectiveness of backdoor attack methods. 
 In our approach, we employ a coefficient $\alpha$ with a small value (i.e., $\alpha$=1.0) to improve the stealthy of triggers during the training process, while a larger value (i.e., $\alpha$=100.0) is used
 to enhance the impact of triggers and further improve the effectiveness of WaveAttack. This method ensures that the backdoored samples during the inference process have sufficient ``power'' to activate the backdoor in DNN, thus achieving a high ASR.

\section{Experiments}\label{experiment}
\label{exps}

To demonstrate the effectiveness and stealthiness of our approach, 
we implemented  WaveAttack  using  Pytorch and  compared the performance of  WaveAttack with
 seven existing backdoor attack methods. 
We conducted all experiments on a workstation with a 3.6GHz Intel i9 CPU, 32GB of memory, an NVIDIA GeForce RTX3090 GPU, and a Ubuntu operating system.
We designed comprehensive experiments to address the following three research questions:

\textbf{RQ1 (Effectiveness of WaveAttack)}: Can WaveAttack successfully inject backdoors into DNNs? 

\textbf{RQ2 (Stealthiness of WaveAttack)}: How do the stealthiness of poisoned samples generated by WaveAttack compare to those generated by state-of-the-art (SOTA) methods?

 \textbf{RQ3 (Resistance to Existing Defenses)}: Can WaveAttack resist existing defense methods?

 \begin{table*}[ht] 
\caption{Attack performance comparison between WaveAttack and seven SOTA attack methods.}
\centering
\footnotesize
\begin{tabular}{c|cc|cc|cc|cc}
\hline
\multirow{2}{*}{Method} & \multicolumn{2}{c|}{CIFAR-10} & \multicolumn{2}{c|}{CIFAR-100} & \multicolumn{2}{c|}{GTSRB} & \multicolumn{2}{c}{ImageNet} \\ \cline{2-9} 
                        & BA            & ASR           & BA             & ASR           & BA          & ASR          & BA     & ASR                        \\ \hline
No attack               & 94.59         & -             & 75.55          & -             &  99.00           & -            &  87.00      & -    \\ \hline
BadNets \cite{badnets}                 & 94.36         & \textbf{100}           & 74.90          & \textbf{100}           &    98.97         &  \textbf{100}            &  85.80      & \textbf{100}                           \\
Blend \cite{physics_backdoor}                   & 94.51         & 99.91         & 75.10          & 99.84         &    98.26         &  \textbf{100}            &   86.40     & \textbf{100}                           \\
IAD \cite{backdoorat-IAD}                     & 94.32         & 99.12         & 75.14          & 99.28         &      99.26       &  98.37            &   -     &   -                         \\
WaNet \cite{backdoorat-WaNet}                   & 94.23         & 99.57         & 73.18          & 98.52         &    99.21         &   99.58           &   86.60     &  89.20                          \\
BppAttack \cite{backdoorat-Bpp}               & 94.10         & \textbf{100}           & 74.68          & \textbf{100}           &     98.93        &    99.91          &   85.90     &  99.50                          \\
Adapt-Blend \cite{backdoorat-adapt}                 & 94.31         & 71.57           & 74.53          & 81.66           &   98.76          &    60.25          &   86.40     &   90.10                         \\
FTrojan \cite{backdoorat-FTrojan}                 & 94.29         & \textbf{100}           & 75.37          & \textbf{100}           &   98.83          &    \textbf{100}          &   85.10     &   \textbf{100}                         \\ \hline
WaveAttack              & \textbf{94.55}         & \textbf{100}         & \textbf{75.17}          & 99.16         &  \textbf{99.30}           &  \textbf{100}            &  \textbf{86.60}      &  97.60                          \\ \hline
\end{tabular}   \label{exp_effectiveness}
\end{table*}

\begin{figure*}[ht]
\centering
\includegraphics[width=5.5in]{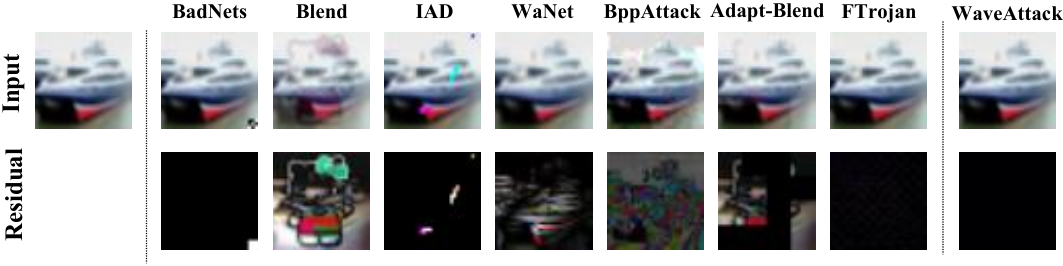}
\vspace{-0.10in}
\caption{Comparison of examples generated by seven backdoor attacks. For each attack, we show the poisoned sample (top) and the magnified ($\times$5) residual (bottom).} \label{figure_trigger}
\vspace{-0.15in}
\end{figure*}

\subsection{Experimental Settings}\label{exp_set}

\textbf{Datasets and DNNs.}
We evaluated all the attack methods on four classical benchmark datasets, i.e.,  CIFAR-10 \cite{krizhevsky2009learning}, CIFAR-100 \cite{krizhevsky2009learning}, GTSRB \cite{stallkamp2012man} and a subset of ImageNet (with the first 20 categories) \cite{deng2009ImageNet}. The statistics of datasets adopted in the experiments are presented in Table~\ref{dataset} (see Appendix~\ref{appendix_dataset}). We used ResNet18 \cite{he2016deep} as the DNN for the experiments. Moreover, we used VGG16 \cite{vgg}, SENet18 \cite{SEresnet}, ResNeXt29 \cite{resnext}, and DenseNet121 \cite{densenet} to evaluate the generalizability of WaveAttack.

\textbf{Attack Configurations.}
To compare the performance of WaveAttack with SOTA attack methods, we considered seven SOTA backdoor attacks, i.e., BadNets \cite{badnets}, Blend \cite{physics_backdoor}, IAD \cite{backdoorat-IAD}, WaNet \cite{backdoorat-WaNet}, BppAttack \cite{backdoorat-Bpp}, Adapt-Blend \cite{backdoorat-adapt}, and FTrojan \cite{backdoorat-FTrojan}. Note that, similar to our work, Adapt-Blend has asymmetric triggers, and FTrojan is also a frequency-based attack. We performed the attack methods using the default hyperparameters described in their original papers. Specifically, the poisoning rate is set to 10\% with a target label of 0 to ensure a fair comparison. See Appendix \ref{appendix_attack} for more details.


\textbf{Evaluation Metrics.}
Similar to the existing work in  \cite{backdoorat-FTrojan}, we evaluated the effectiveness of all attack methods using two metrics, i.e., Attack Success Rate (ASR) and Benign Accuracy (BA). To evaluate the stealthiness of all attack methods, we used three metrics, i.e., Peak Signal-to-Noise Ratio (PSNR) \cite{huynh2008scope}, Structure Similarity Index Measure (SSIM) \cite{wang2004image}, and Inception Score (IS) \cite{salimans2016improved}.

\subsection{Effectiveness Evaluation (RQ1)} 
\label{section_effectiveness}

\textbf{Effectiveness Comparison with SOTA Attack Methods.} To evaluate the effectiveness of WaveAttack, we compared the ASR and BA of WaveAttack with seven SOTA attack methods. Since the IAD \cite{backdoorat-IAD} cannot attack the ImageNet dataset based on their open-source code, we do not provide its comparison result. Table \ref{exp_effectiveness} shows the attack performance of different attack methods. From this table, we can find that WaveAttack can acquire the high ASR without degrading BA obviously. Especially, for the dataset CIFAR-10 and GTSRB, WaveAttack achieves the best ASR and BA than other SOTA attack methods. 
Compared to the FTrojan, a frequency-based attack method, WaveAttack outperforms FTrojan in BA for the datasets CIFAR-10, GTSRB, and ImageNet. 
Note that compared to the asymmetric-based method Adapt-Blend, WaveAttack can obtain superior performance in terms of ASR and BA for all datasets. 

\textbf{Effectiveness on Different Networks.} To evaluate the effectiveness of WaveAttack on various networks, we conducted experiments on CIFAR-10 using different networks (i.e., VGG16 \cite{vgg}, SENet18 \cite{SEresnet}, ResNeXt29 \cite{resnext}, and DenseNet121 \cite{densenet}). Table \ref{exp_other_networks} shows the attack performance of WaveAttack on these networks. From this table, we can find that our approach WaveAttack can successfully embed the backdoor into different networks. WaveAttack can not only cause malicious impacts of backdoor attacks, but also maintain a classification performance with a high BA, demonstrating the generalizability of WaveAttack.

\begin{table}[ht]
\vspace{-0.05in}
\caption{Effectiveness on different DNNs.}
\vspace{-0.10in}
\centering
\footnotesize
\begin{tabular}{c|c|cc}
\hline
\multirow{2}{*}{Network} & No Attack & \multicolumn{2}{c}{WaveAttack} \\ \cline{2-4} 
                         & BA        & BA             & ASR           \\ \hline
VGG16                    & 93.62     & 93.70          & 99.76         \\
SENet18                  & 94.51     & 94.63          & 100           \\
ResNeXt29                & 94.79     & 95.08          & 100           \\
DenseNet121               & 95.29     & 95.10          & 99.78         \\ \hline
\end{tabular}   \label{exp_other_networks}
\vspace{-0.15in}
\end{table}

\begin{table*}[ht]
\caption{Stealthiness comparison with existing attacks. Larger PSNR, SSIM, and smaller IS indicate better performance. The best and the second-best results are \textbf{highlighted} and \underline{underlined}, respectively.}
\centering
\footnotesize
\setlength{\tabcolsep}{0.85mm}{
\begin{tabular}{c|ccc|ccc|ccc|ccc}
\hline
\multirow{2}{*}{Attack Method} & \multicolumn{3}{c|}{CIFAR-10}                                                                                 & \multicolumn{3}{c|}{CIFAR-100}                                                                                & \multicolumn{3}{c|}{GTSRB}                                                                                    & \multicolumn{3}{c}{ImageNet} \\ \cline{2-13} 
                               & PSNR                               & SSIM                                & IS                                 & PSNR                               & SSIM                                & IS                                 & PSNR                               & SSIM                                & IS                                 & PSNR     & SSIM     & IS      \\ \hline
No Attack                      & INF                                & 1.0000                              & 0.000                              & INF                                & 1.0000                              & 0.000                              & INF                                & 1.0000                              & 0.000                              & INF      & 1.0000   & 0.000   \\ \hline
BadNets \cite{badnets}                        & 25.77                              & 0.9942                              & 0.136                              & 25.48                              & 0.9943                              & 0.137                              & 25.33                              & \textbf{0.9935}    & 0.180                              & 21.88    & 0.9678   & 0.025   \\
Blend \cite{physics_backdoor}                          & 20.40                              & 0.8181                              & 1.823                              & 20.37                              & 0.8031                              & 1.60                               & 18.58                              & 0.6840                              & 2.118                              & 13.72    & 0.1871   & 2.252   \\
IAD \cite{backdoorat-IAD}                           & 24.35                              & 0.9180                              & 0.472                              & 23.98                              & 0.9138                              & 0.490                              & 23.84                              & 0.9404                              & 0.309                              & -        & -        & -       \\
WaNet \cite{backdoorat-WaNet}                         & 30.91                              & 0.9724                              & 0.326                              & 31.62                              & 0.9762                              & 0.237                              & 33.26                              & 0.9659                              & 0.170                              & 35.18    & \underline{0.9756}   & 0.029   \\
BppAttack \cite{backdoorat-Bpp}                     & 27.79                              & 0.9285                              & 0.895                              & 27.93                              & 0.9207                              & 0.779                              & 27.79                              & 0.8462                              & 0.714                              & 27.34    & 0.8009   & 0.273   \\
Adapt-Blend \cite{backdoorat-adapt}                   & 25.97                              & 0.9231                              & 0.519                              & 26.00                              & 0.9133                              & 0.495                              & 24.14                              & 0.8103                              & 1.136                              & 18.96    & 0.6065   & 1.150   \\
FTrojan \cite{backdoorat-FTrojan}                       & \underline{44.07} & \underline{0.9976} & \underline{0.019} & \underline{44.09} & \underline{0.9972} & \underline{0.017} & \underline{40.23} & 0.9813                              & \underline{0.065} & \underline{35.55}    & 0.9440   & \underline{0.013}   \\ \hline
WaveAttack                     & \textbf{47.49}    & \textbf{0.9979}    & \textbf{0.011}    & \textbf{50.12}    & \textbf{0.9992}    & \textbf{0.005}    & \textbf{40.67}    & \underline{0.9877} & \textbf{0.058}    & \textbf{45.60}    & \textbf{0.9913}   & \textbf{0.007}   \\ \hline
\end{tabular}}   \label{table_stealthiness}
\vspace{-0.1in}
\end{table*}

\subsection{Stealthiness Evaluation (RQ2)} 
\label{section_stealthiness}
To evaluate the stealthiness of WaveAttack, we compared the images with triggers generated by WaveAttack with the ones of SOTA attack methods. Moreover, we used t-SNE \cite{van2008visualizing} to visualize the latent spaces for poisoned samples and benign samples from the target label.

\textbf{Stealthiness Results from The Perspective of Images.}
To show the stealthiness of triggers generated by WaveAttack, Figure \ref{figure_trigger} compares WaveAttack and SOTA attack methods using poisoned samples and their magnified residuals ($\times$5) counterparts. From this figure, we can find that the residual generated by WaveAttack is the smallest and leaves only a few subtle artifacts. The injected trigger by WaveAttack is nearly invisible to humans.

We used three metrics (i.e., PSNR, SSIM, and IS) to evaluate the stealthiness of triggers generated by WaveAttack. Table \ref{table_stealthiness} shows the stealthiness comparison results between WaveAttack and seven SOTA attack methods. From this table, we can find that for datasets CIFAR-10, CIFAR-100, and ImageNet, WaveAttack achieves the best stealthiness. Note that WaveAttack can achieve the second-best SSIM for dataset GTSRB, but it outperforms BadNets by up to 60.56\% in PSNR and 67.5\% in IS.

\begin{figure}[h]
 \vspace{-0.10in}
  \centering
  \subfigure[BadNets\label{featureSpace:Badnets}]{\includegraphics[width=0.98in]{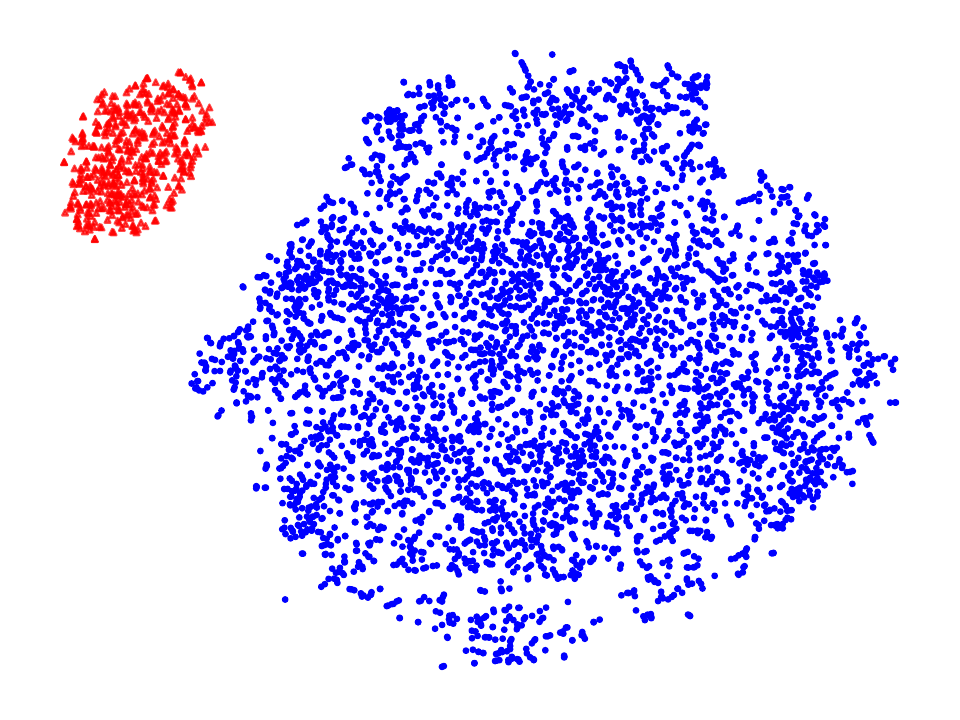}}
  \subfigure[Blend\label{featureSpace:Blend}]{\includegraphics[width=0.98in]{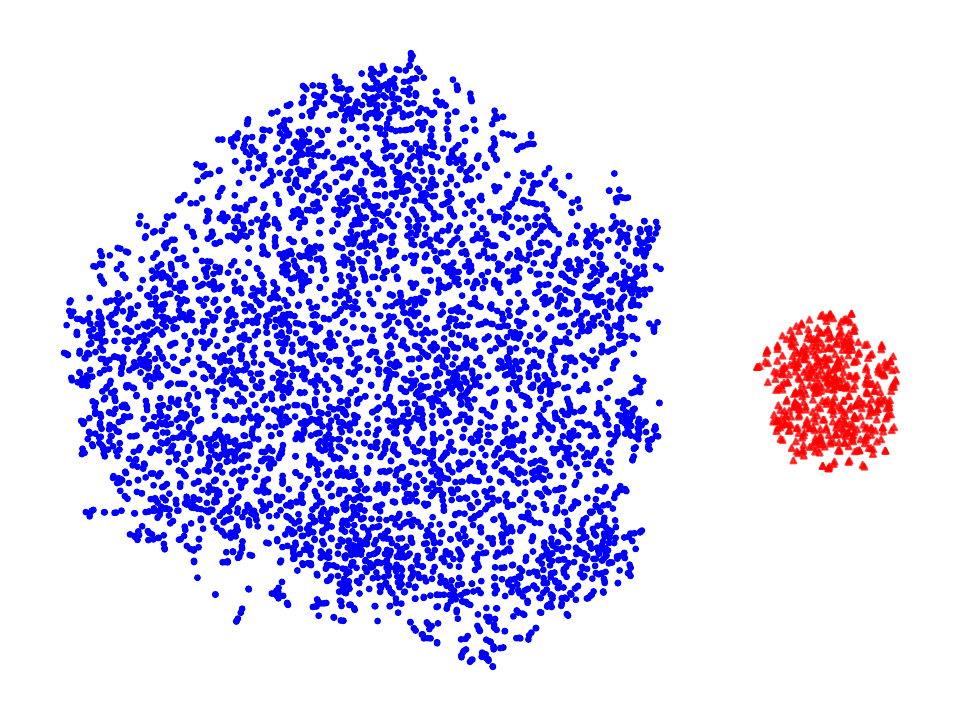}}
   \subfigure[WaNet\label{featureSpace:WaNet}]{\includegraphics[width=0.98in]{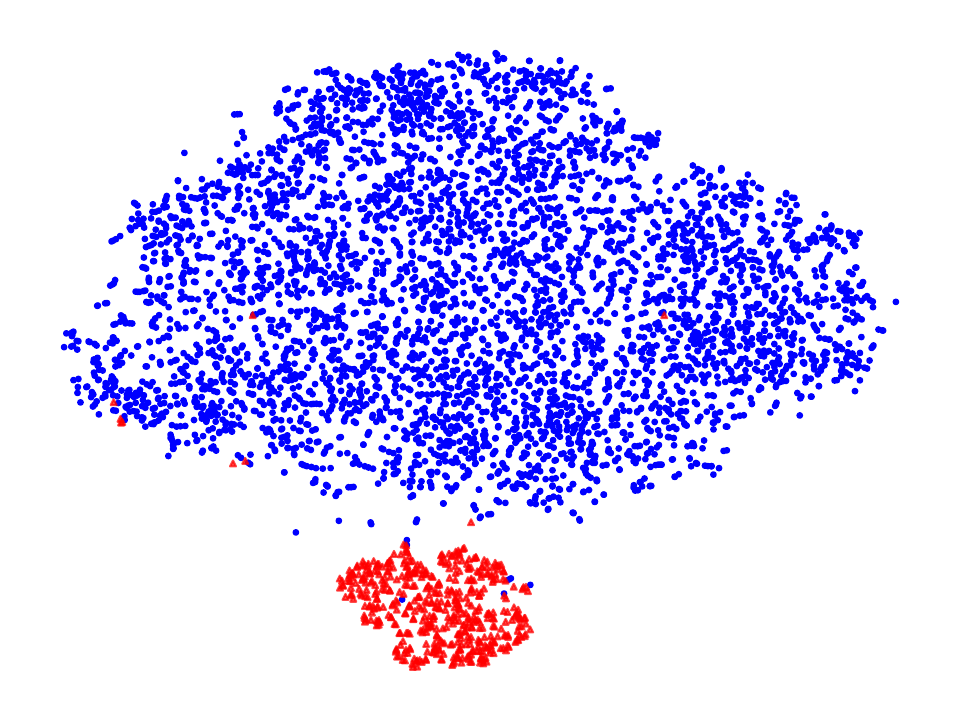}}
   \subfigure[Adapt-Blend\label{featureSpace:adaptBlend}]{\includegraphics[width=0.98in]{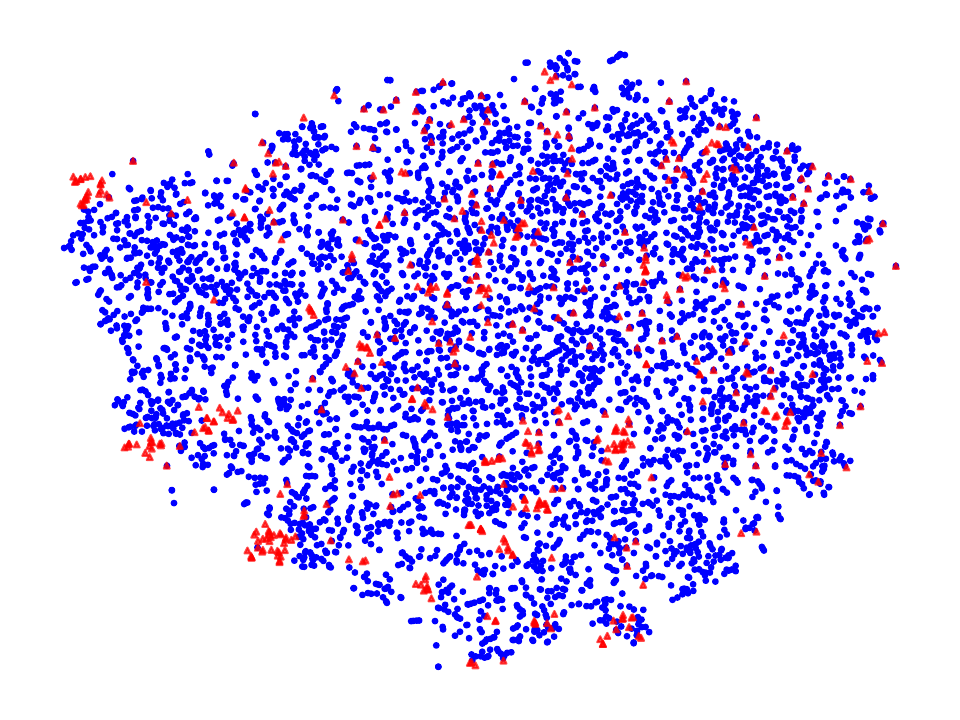}} 
   \subfigure[FTrojan\label{featureSpace:FTrojan}]{\includegraphics[width=0.98in]{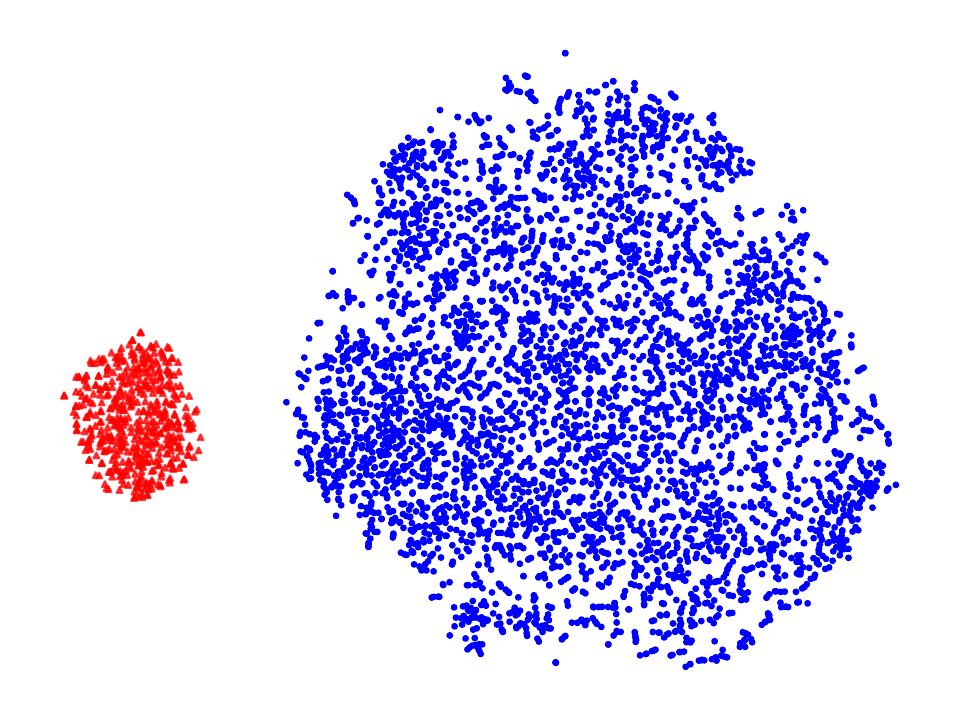}} 
   \subfigure[WaveAttack\label{featureSpace:WaveAttack}]{\includegraphics[width=0.98in]{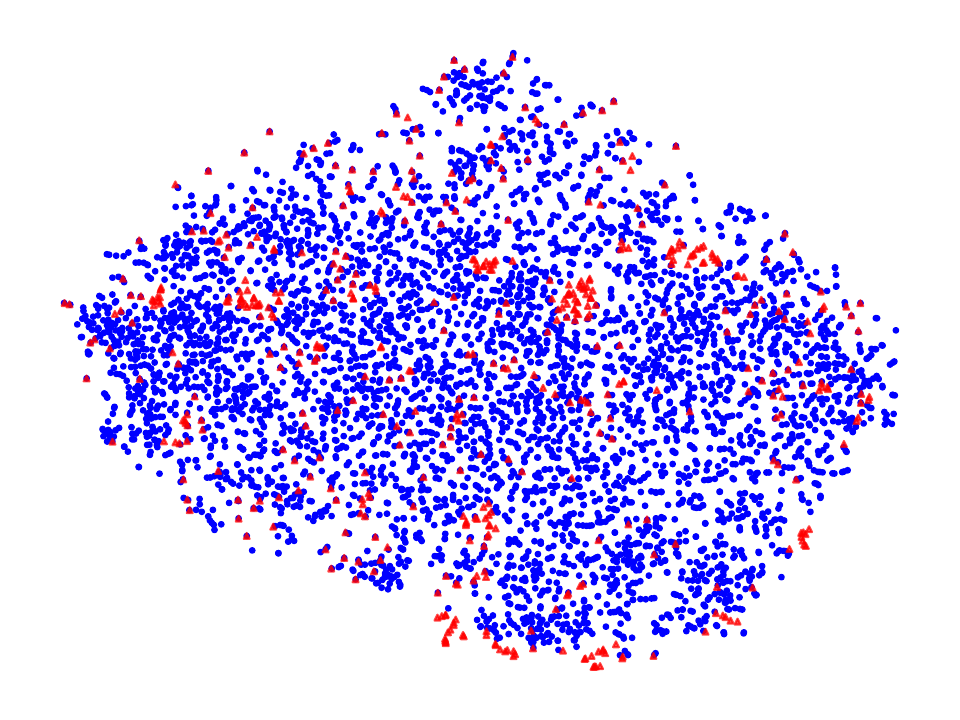}} \\
   \vspace{-0.1in}
  \caption{The t-SNE of feature vectors in the latent space under different attacks on CIFAR-10. We use \textcolor{red}{red} and \textcolor{blue}{blue} points to denote \textcolor{red}{poisoned } and  \textcolor{blue}{benign samples}, respectively, where each point in the plots corresponds to a training sample from the target label.} \label{figure_feature_space}
\vspace{-0.15in}
\end{figure}

\textbf{Stealthiness Results from The Perspective of Latent Space.}
There are so many backdoor defense methods \cite{tran2018spectral, hayase2021spectre} based on the assumption that there is a latent separation between poisoned and benign samples in latent space. Therefore, ensuring the stealthiness of the attack method from the perspective of the latent space becomes necessary.
We obtained feature vectors of the test result from the feature extractor (the DNN without the last 
classifier layer) and used t-SNE \cite{van2008visualizing} for visualization.
Figure \ref{figure_feature_space} visualizes the distributions of feature representations of the poisoned samples and the benign samples from the target label under the six attacks. From Figure \labelcref{featureSpace:Badnets,featureSpace:Blend,featureSpace:WaNet,featureSpace:FTrojan}, we can observe that there are two distinct clusters, which can be utilized to detect poisoned samples or backdoored models \cite{backdoorat-adapt}. However, as shown in \labelcref{featureSpace:adaptBlend,featureSpace:WaveAttack}, we can find that the feature representations of poisoned samples are intermingled with those of benign samples for Adapt-Blend and WaveAttack, i.e., there is only one cluster. Adapt-Blend and WaveAttack can achieve the best stealthiness from the perspective of latent space and break the latent separation assumption to evade backdoor defenses. Although Adapt-Blend exhibits a degree of stealthiness, Table \ref{table_stealthiness} reveals that WaveAttack surpasses Adapt-Blend in terms of image quality, thus suggesting that WaveAttack can achieve superior stealthiness.

\subsection{Resistance to Existing Defenses (RQ3)}
To evaluate the robustness of WaveAttack against existing backdoor defenses, we implemented representative backdoor defenses (i.e., GradCAM \cite{DBLP:journals/ijcv/SelvarajuCDVPB20}, STRIP \cite{DBLP:conf/acsac/GaoXW0RN19}, Fine-Pruning \cite{liu2018fine}, and Neural Cleanse \cite{NC}) and evaluated the resistance to them. We also show the robustness of WaveAttack against Spectral Signature \cite{tran2018spectral} in the appendix.

\begin{figure}[h]
 \vspace{-0.15in}
\centering
\includegraphics[width=2.8in]{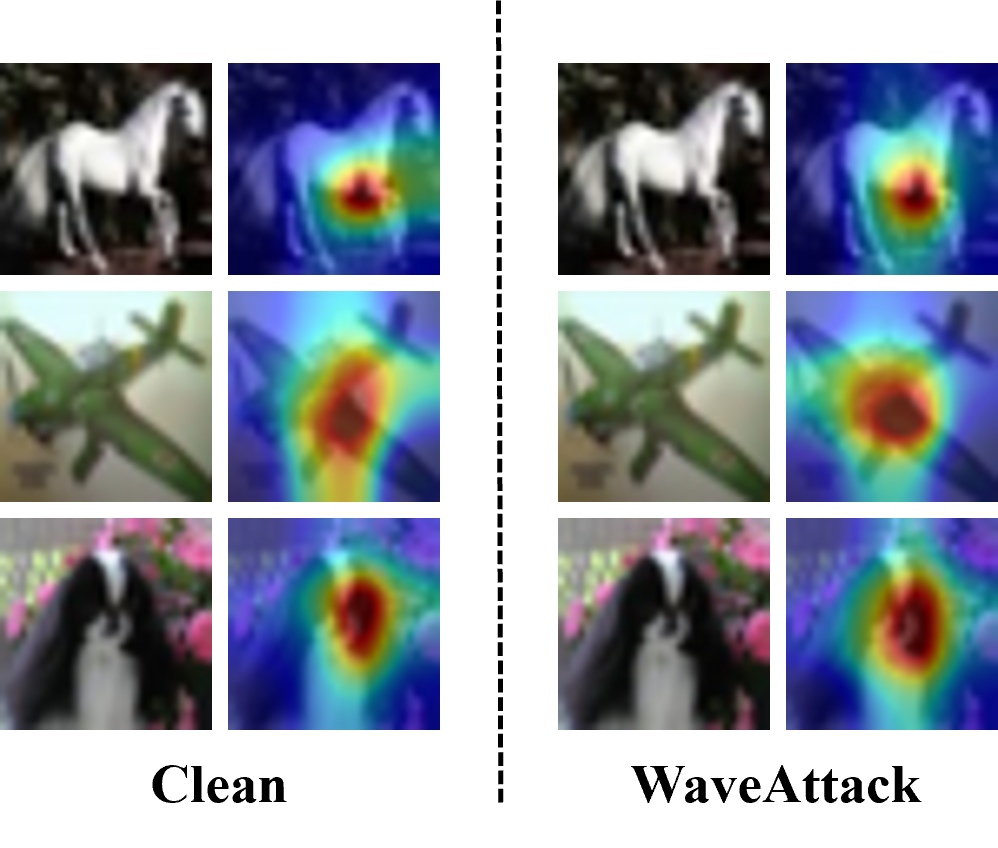}
\vspace{-0.10in}
\caption{GradCAM visualization results for both  clean  and backdoored models.} \label{figure_GradCAM}
\vspace{-0.10in}
\end{figure}

\textbf{GradCAM}. 
As an effective visualizing mechanism,  GradCAM \cite{DBLP:journals/ijcv/SelvarajuCDVPB20} has been used to visualize  intermediate feature maps of DNN,  interpreting  the DNN predictions.
Existing defense methods \cite{DBLP:journals/corr/abs-1812-00292,doan2020februus} exploit GradCAM 
to analyze the heatmap of input samples. 
Specifically, a clean model correctly predicts the class label, whereas a backdoored model predicts the target label. 
Based on this phenomenon, the backdoored model can induce an abnormal GradCAM heatmap compared with the clean model. 
If the heatmaps of poisoned samples are similar to those of benign sample counterparts, the attack method is robust and can resist defense methods based on GradCAM. Figure \ref{figure_GradCAM} shows the visualization heatmaps of a clean model and a backdoored model attacked by WaveAttack.
Please note that here ``clean'' denotes a clean model trained by using benign training datasets.
From this figure, we can find that the heatmaps of these models are similar, and WaveAttack can resist defense methods based on GradCAM. 

\begin{figure}[h]
  \centering
  \subfigure[CIFAR-10\label{strip:cifar10}]{\includegraphics[width=0.325\linewidth]{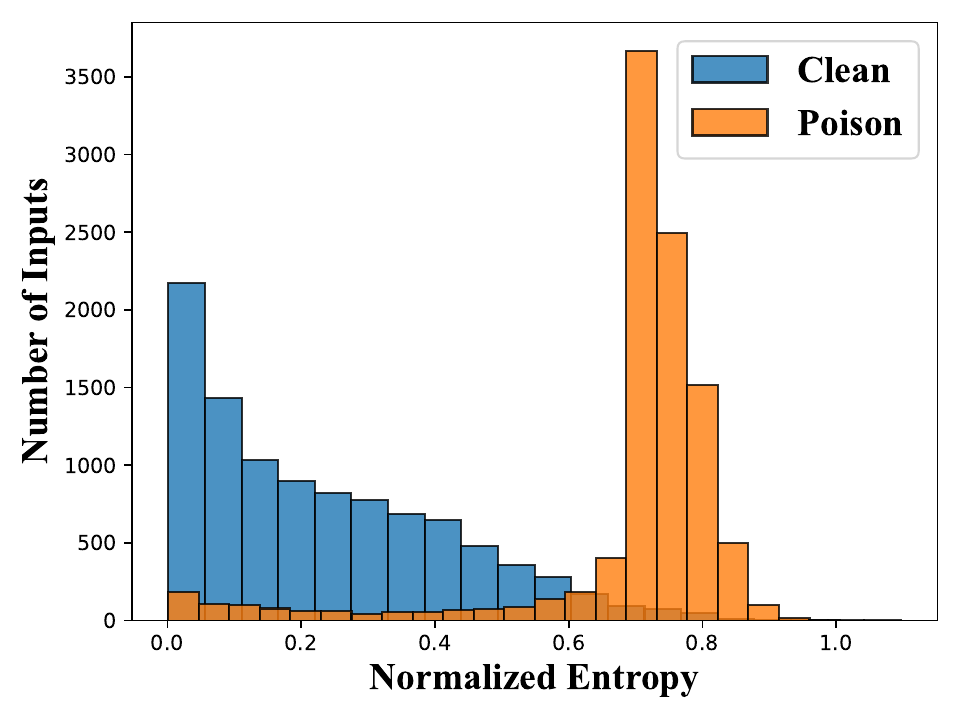}}
  \hspace{-0.05in}
  \subfigure[CIFAR-100\label{strip:cifar100}]{\includegraphics[width=0.32\linewidth]{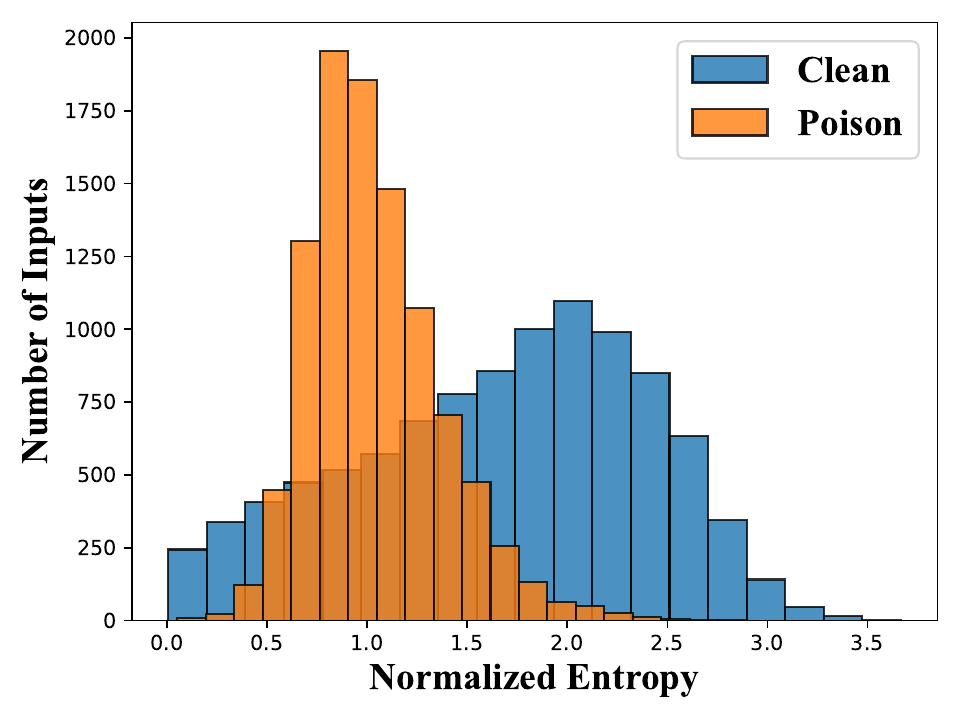}}
  \hspace{-0.02in}\subfigure[GTSRB\label{strip:gtsrb}]{\includegraphics[width=0.32\linewidth]{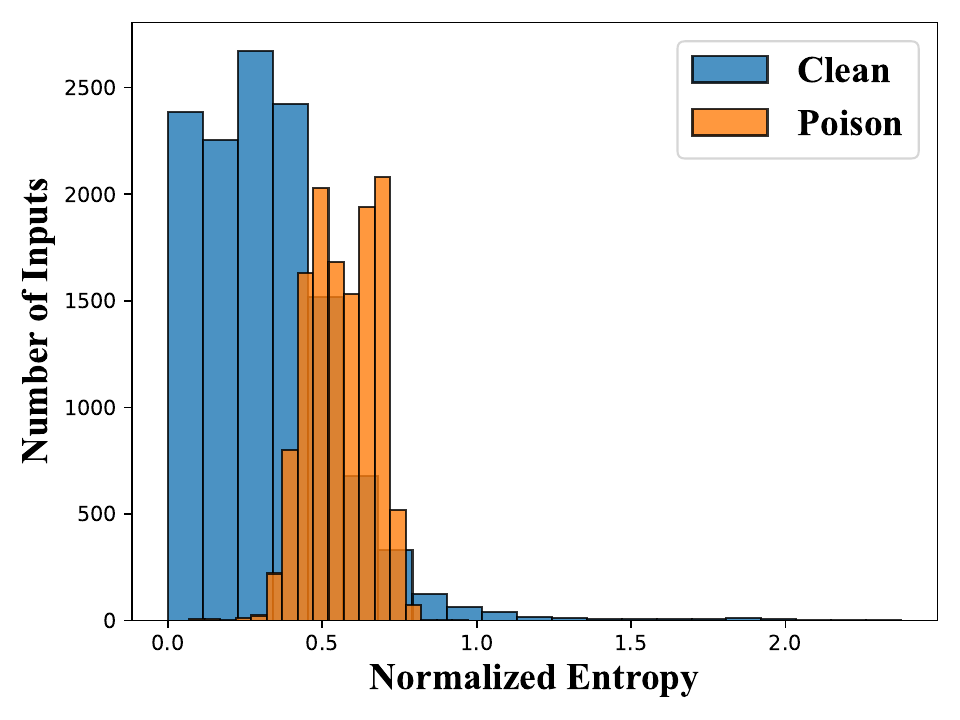}} 
  \vspace{-0.10in}
  \caption{STRIP normalized entropy  of WaveAttack.} \label{figure_strip}
\end{figure}

\textbf{STRIP}. STRIP \cite{DBLP:conf/acsac/GaoXW0RN19} is a representative sample-based defense method. When inputting a  sample potentially poisoned to a model, STRIP will perturb it through a random set of clean samples and monitor the entropy of the prediction output. If the entropy of an input sample is low, STRIP will consider it poisoned.
Figure \ref{figure_strip} shows the entropies of the benign and poisoned samples. 
From this figure, we can see the entropies of the poisoned samples are bigger than those of the benign samples, and STRIP fails to detect the poisoned samples generated by WaveAttack.

\begin{figure}[h]
\centering
\includegraphics[width=3.4in]{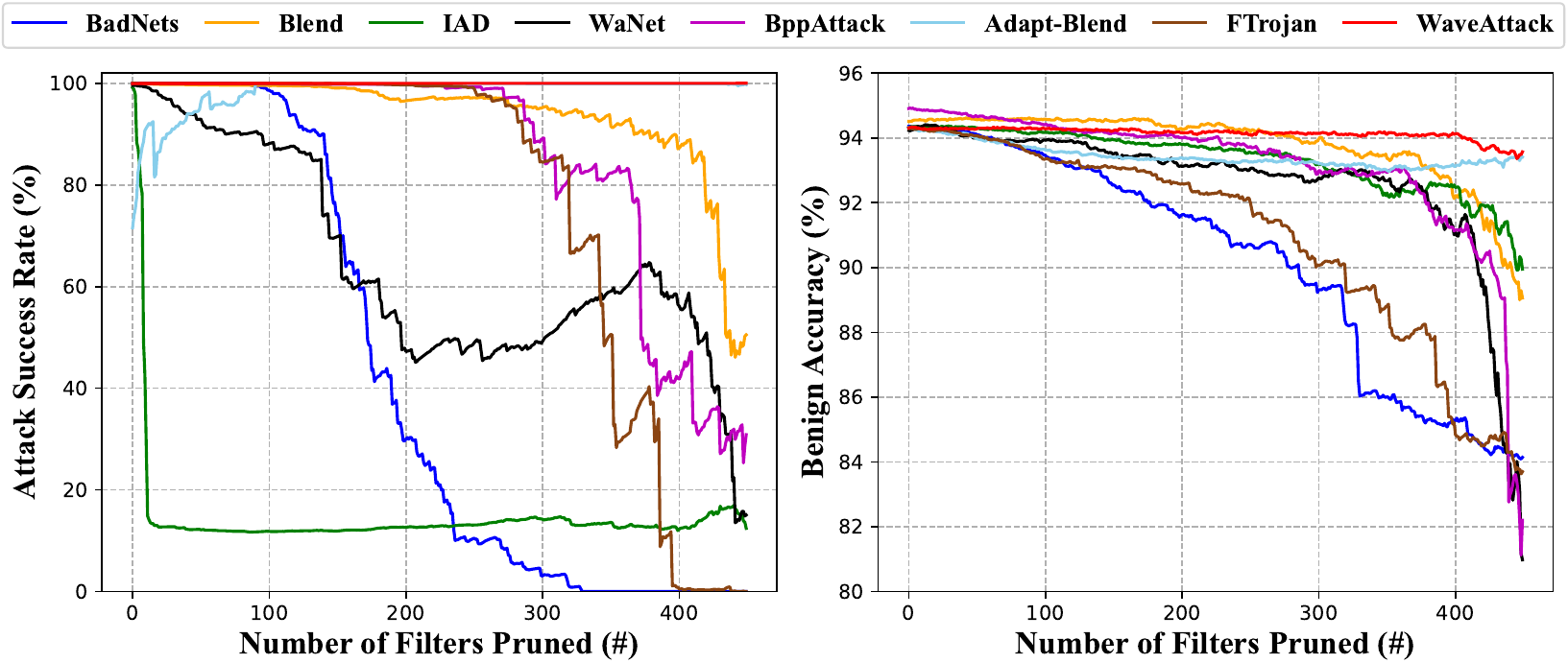}
\caption{ASR  comparison  against Fine-Pruning.} \label{figure_pruning}
\end{figure}

\textbf{Fine-Pruning}. Fine-Pruning (FP) \cite{liu2018fine} is a representative model reconstruction defense, which is based on the assumption that the backdoor can activate a few dormant neurons in DNNs. 
Therefore, pruning these dormant neurons can eliminate the backdoors in DNNs. 
To evaluate the resistance to FP, we gradually pruned the neurons of the last convolutional and fully-connected layers.
Figure \ref{figure_pruning} shows the performance comparison between WaveAttack and seven SOTA attack methods on CIFAR-10 by resisting FP.
From this figure, we can find that along with the more neurons are pruned,  WaveAttack can acquire superior performance than other SOTA attack methods in terms of both ASR and BA. 
In other words, Fine-Pruning is not able to  eliminate the backdoor generated by WaveAttack. Note that, though the ASR and BA of WaveAttack are similar to those of Adapt-Blend at the final  stage of pruning, the initial ASR of Adapt-Blend (i.e., 71.57\%) is much lower than that of WaveAttack (i.e., 100\%).

\begin{figure}[h]
\centering
\includegraphics[width=2.1in]{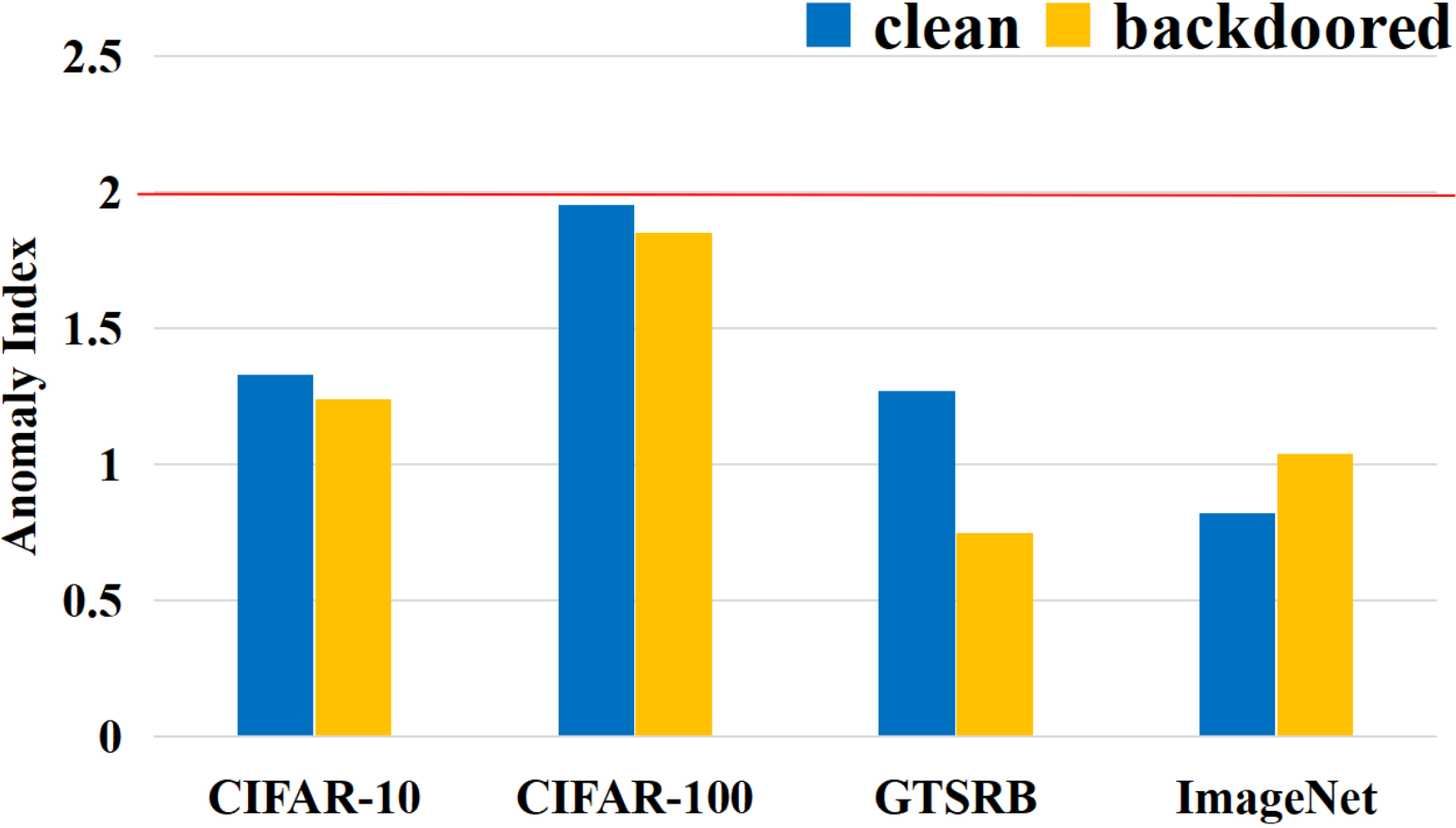}
\caption{Defense performance against NC.} \label{exp_nc}
 \vspace{-0.15in}
\end{figure}

\textbf{Neural Cleanse}. As a representative trigger generation defense method, Neural Cleanse (NC) \cite{NC} assumes that the trigger designed by the adversary is small. 
Initially, NC optimizes a trigger pattern for each class label via an optimization process. 
Then, NC uses Anomaly Index (i.e., Median Absolute Deviation \cite{MAD}) to detect whether a DNN is backdoored. 
Similar to the work \cite{NC}, we think the DNN is backdoored if the anomaly index is larger than 2. 
To evaluate the resistance to NC, we conducted experiments to evaluate our approach WaveAttack by resisting NC. 
Figure \ref{exp_nc} shows the defense results against NC. Please note that here ``clean'' denotes clean models trained by using benign training datasets, and ``backdoored'' denotes backdoored models by WaveAttack that are from the  Subsection \ref{section_effectiveness}. 
From this figure, we can find that the abnormal index of WaveAttack is smaller than 2 for all datasets, and WaveAttack can bypass the NC detection.

\section{Conclusion}\label{conclusion}

Although backdoor attacks on DNNs have attracted increasing attention from adversaries, few of them consider both fidelity of poisoned samples and latent space simultaneously to enhance the stealthiness of their attack methods.
To establish an effective and stealthy backdoor attack against various backdoor detection techniques, this paper proposed a novel frequency-based method named WaveAttack, which employs DWT to extract high-frequency features from samples for backdoor trigger generation.
Based on our proposed frequency obfuscation method, WaveAttack can maintain high effectiveness and stealthiness, thus remaining undetectable by both human inspection and backdoor detection mechanisms. 
Furthermore, we introduced an asymmetric frequency obfuscation method to improve the impact of triggers and further enhance the effectiveness of WaveAttack. 
Comprehensive experimental results show that, compared with various SOTA backdoor attack methods, WaveAttack can not only achieve both higher stealthiness and effectiveness but also minimize the impact of image quality on four well-known datasets.

\clearpage

\clearpage
\appendix

\section{Implementation Details for Experiments}     \label{appendix_main_exp}

\subsection{Settings of Datasets}   
\label{appendix_dataset}
Table \ref{dataset} presents the setting of datasets used in our experiments.
\begin{table}[ht]
\caption{Datasets Settings.}
\footnotesize
\centering
\setlength{\tabcolsep}{0.85mm}{
\begin{tabular}{c c c c c}
\hline
Dataset  & Input Size  & Classes & \tabincell{c}{Training\\Images} & \tabincell{c}{Test\\Images}\\ \hline \hline
CIFAR-10 & 3$\times$32$\times$32 & 10      & 50000            & 10000       \\ 
CIFAR-100 & 3$\times$32$\times$32 & 100      & 50000            & 10000       \\ 
GTSRB    & 3$\times$32$\times$32 & 43      & 26640            & 12569       \\
ImageNet subset    & 3$\times$224$\times$ 224 & 20      & 26000            & 1000    \\ \hline
\end{tabular}} \label{dataset}
\end{table}

\subsection{Settings of Attacks} \label{appendix_attack}
For a fair comparison, the settings of WaveAttack are consistent with those of the other seven SOTA attack methods. 
We used the SGD optimizer for training a classifier with a learning rate of 0.01, and the Adam optimizer for training a generator with a learning rate of 0.001. 
We decreased this learning rate by a factor of 10 after every 100 epochs.
We considered various data augmentations, i.e., random crop and random horizontal flipping. 
For BadNets, we used a grid trigger placed in the bottom right corner of the image. 
For Blend, we applied a ``Hello Kitty''  trigger on CIFAR-10, CIFAR-100, and GTSRB datasets and used random noises on the ImageNet dataset. 
For other attack methods, we used the default settings in their respective papers.

\subsection{Resistance to Spectral Signature.} \label{exp_spectral}
\begin{figure}[h]
  \centering
  \subfigure[CIFAR-10\label{spectral:cifar10}]{\includegraphics[width=2.4in]{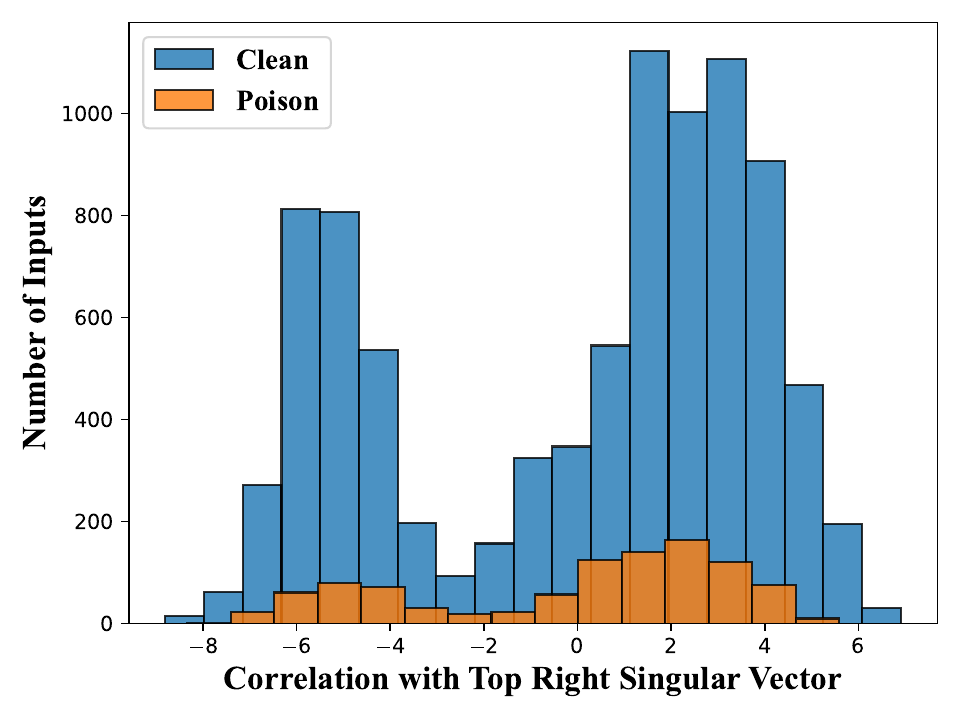}}
  \subfigure[CIFAR-100\label{spectral:cifar100}]{\includegraphics[width=2.4in]{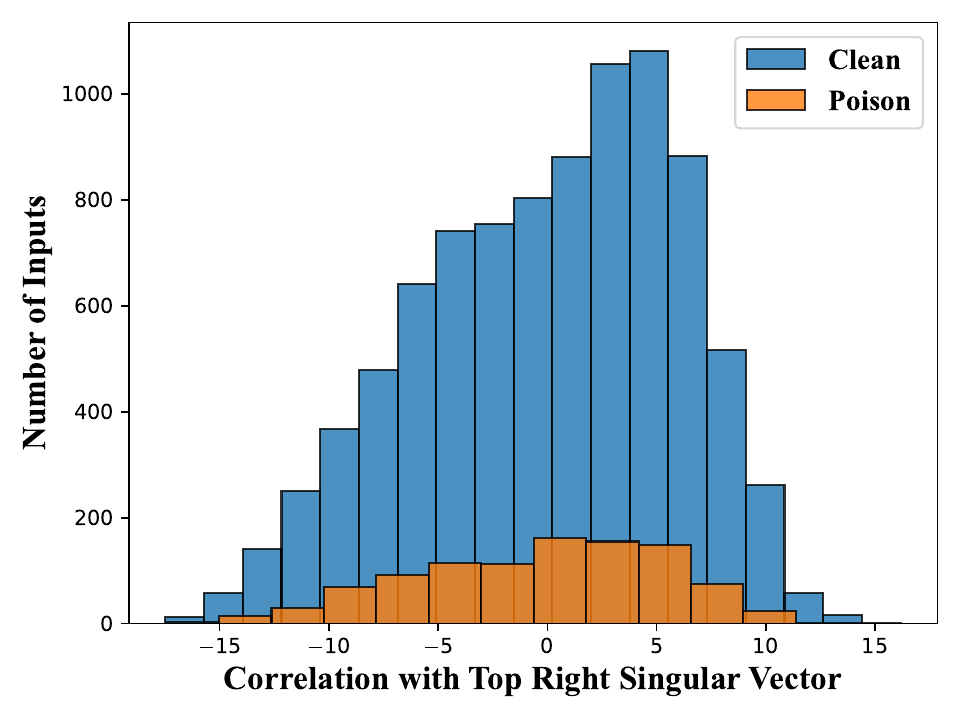}}
  \subfigure[GTSRB\label{spectral:gtsrb}]{\includegraphics[width=2.4in]{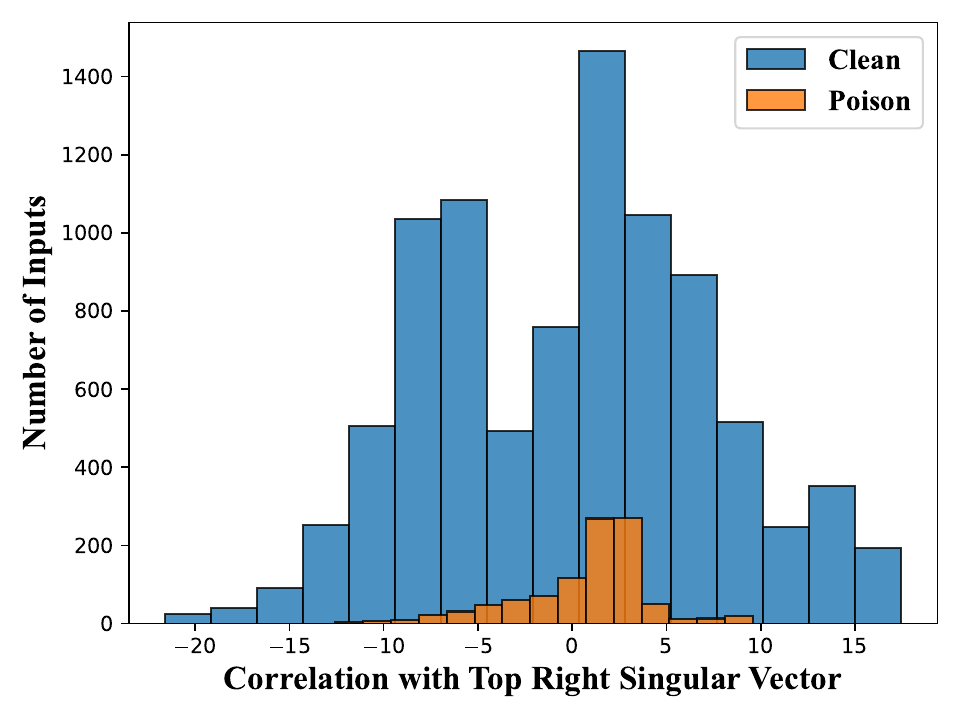}} 
    \vspace{-0.15in}
  \caption{The correlation with top right singular vector on different datasets.} \label{figure_spectral}
  \vspace{-0.2in}
\end{figure}

Spectral Signature \cite{tran2018spectral} is a representative latent space-based detection defense method.  Given a set of benign and poisoned samples, Spectral Signature first collects their latent features and computes the top singular value of the covariance matrix. Then, for each sample, it calculates the correlation score between its features and the top singular value used as the outlier score. If the samples with high outlier scores, they will be evaluated as poisoned samples. We randomly selected 9000 benign samples and 1000 poisoned samples. Figure \ref{figure_spectral} shows histograms of the correlations between latent features of samples and the top right singular vector of the covariance matrix. 
From this figure, we can see that the histograms of the poison data are similar to those of the benign data. Therefore, Spectral Signature fails to detect the poison data generated by WaveAttack.

\subsection{Broader Impact and Limitations}\label{limitation}
\textbf{Broader Impact.} In this work, we introduce a new effective and stealthy backdoor attack method named WaveAttack, which can stealthily compromise security-critical systems. 
If used improperly, the proposed attack method may pose a security risk to the existing DNN applications. Nevertheless, we hope that by emphasizing the potential harm of this malicious threat model, our work will stimulate the development of stronger defenses and promote greater attention from experts in the field.
As a result, this knowledge  promotes the creation of more secure and dependable DNN models and robust defensive measures.

\textbf{Limitations.}  WaveAttack requires more computing resources and runtime overhead than most existing backdoor attack methods, due to the necessity of training a generator $g$ to generate residuals of the high-frequency component. 
Moreover, we do not consider a more standard threat model, in which the adversary can only control the training dataset. 
In this threat model, we used our pre-trained generator to modify some benign samples in the training dataset. Our approach WaveAttack has limited effectiveness. This limitation also appears in Qi et al. \cite{backdoorat-adapt}. In the future, we will explore more effective and stealthy backdoor attack methods under this threat model.



\end{document}